\documentclass[sigconf]{acmart}
\AtBeginDocument{%
	\providecommand\BibTeX{{%
			\normalfont B\kern-0.5em{\scshape i\kern-0.25em b}\kern-0.8em\TeX}}}

\usepackage[linesnumbered,ruled]{algorithm2e}
\usepackage[utf8]{inputenc}
\usepackage{enumitem}
\usepackage{rotating}
\usepackage{amsmath}
\usepackage{array}
\usepackage{longtable}
\usepackage{caption}
\usepackage{mathtools}  
\usepackage{tabulary}
\usepackage{subcaption,booktabs}
\usepackage{float}
\usepackage{graphicx}
\usepackage{xcolor}
\usepackage[normalem]{ulem}
\setcopyright{none} 
\renewcommand\footnotetextcopyrightpermission[1]{} 
\begin{document}
\title{Robust Multimodal Fusion for Human Activity Recognition}
\author{Sanju Xaviar}
\email{xaviar@ualberta.ca}
\affiliation{%
  \institution{University of Alberta}
  \city{Edmonton}
  \state{AB}
  \country{Canada}
}
\author{Xin Yang}
\email{xyang18@ualberta.ca }
\affiliation{%
  \institution{University of Alberta}
  \city{Edmonton}
  \state{AB}
  \country{Canada}
}

\author{Omid Ardakanian}
\email{oardakan@ualberta.ca}
\affiliation{%
  \institution{University of Alberta}
  \city{Edmonton}
  \state{AB}
  \country{Canada}
}
\begin{abstract}
The proliferation of IoT and mobile devices equipped with heterogeneous sensors
has enabled new applications that rely on the fusion of time-series data generated by 
multiple sensors with different modalities.
While there are promising deep neural network architectures for multimodal fusion, 
their performance falls apart quickly in the presence of 
consecutive missing data and noise across multiple modalities/sensors, 
the issues that are prevalent in real-world settings.
We propose Centaur, a multimodal fusion model for human activity recognition~(HAR) 
that is robust to these data quality issues.
Centaur combines a data cleaning module, which is a denoising autoencoder with convolutional layers,
and a multimodal fusion module, which is a deep convolutional neural network 
with the self-attention mechanism to capture cross-sensor correlation.
We train Centaur using a stochastic data corruption scheme and evaluate it on three datasets
that contain data generated by multiple inertial measurement units.
We show that Centaur's data cleaning module outperforms two state-of-the-art autoencoder-based architectures, 
and its multimodal fusion module outperforms four strong baselines.
Compared to two robust fusion architectures from the related work, 
Centaur is more robust especially to consecutive missing data that occur in multiple sensor channels,
achieving on average 11.59--17.52\% higher accuracy in the HAR task.
\end{abstract} 



\maketitle
\pagestyle{plain}  

\section{Introduction}


The demand for Internet of Things (IoT) and mobile devices 
has witnessed a steady growth in the last decade.
According to a recent survey~\cite{Deloitee-pr}, US households have on average 22 
entertainment and smart home devices.
Moreover, many people carry multiple mobile and wearable devices, \emph{e.g.}, 
they have a smartphone in their pocket and wear a smartwatch.
These IoT and mobile devices are equipped with a variety of sensors.
One example is the inertial measurement unit~(IMU) embedded in smartphones and smartwatches.
It combines a tri-axial accelerometer, a tri-axial gyroscope, 
and sometimes a tri-axial magnetometer in a system-in-package 
to measure specific force, angular rate, and magnetic field,
enabling applications such as fitness tracking~\cite{Jeya2019}. 
Fusing multimodal data collected by sensors embedded in one or multiple such devices 
helps capture complementary information across different modalities~\cite{xue2019deepfusion}, 
thereby reducing the overall uncertainty and making possible a more comprehensive understanding of 
human activities, health conditions, and hand gestures.

Sensor data streams are intermittent and noisy in real-world settings.
This is primarily because sensors are used in various conditions and environments 
without (re)calibration and proper protection, 
which makes them susceptible to offsets and drifts~\cite{Marathe21}, 
in addition to dislocation, deformation, occlusion, and dirt/dust buildup~\cite{PIRMedic}.
For example, while the total offset and scaling error of most IMUs, 
including LSM9DS1 manufactured by STMicroelectronics and BNO055 by Bosch Sensortec, 
is within 1\%, this error will be much higher 
if the sensor is not dynamically calibrated in the environment.
Moreover, wireless sensors often send data to a node 
that has enough compute power to run the fusion model.
For example, data generated by Vicon's Blue Trident or Xsens's MTw IMUs 
worn on the chest, wrist, and ankle are transferred to a smartphone or computer
where they are fused to detect the activity.
In this case, network problems might cause 
consecutive missing data points in all channels of some sensors.
Lastly, battery-powered sensors typically enter a low-energy state 
when the energy stored in their battery is not sufficient for their operation~\cite{Jackson19}.
This could result in consecutive missing data points in some channels of a sensor
until the battery is recharged.


Noise and missing data impose a significant challenge for the effective fusion of data 
from multiple sensors with different modalities. This is due to two main reasons.
First, existing multimodal deep learning techniques are not designed 
to handle time-varying noise and consecutive missing data~\cite{Ngiam11}.
Augmenting the multimodal fusion models to simultaneously address 
these issues and make desired inferences with high accuracy 
could result in complex architectures that are difficult to train and do not generalize well. 
Second, it is hard to capture complementary information 
from different sensors or modalities when data is incomplete.
Simple imputation techniques, such as zero/mean filling and linear interpolation,
may affect the cross-sensor correlation and lower the inference accuracy.
As a result, most related work that considers data quality issues
handles either missing data~\cite{Xiang2013, Tran2017MissingMI, du2018semi, Yao2018, chen2020hgmf}
or noisy measurements~\cite{Yao2016, Yao2019, Gong2021} only. 
To our knowledge, UniTS~\cite{li2021units} is presently the only multimodal fusion model 
that is designed to be robust to both consecutive missing data and noise.
To train this model, the authors add Gaussian noise to training data, 
and randomly mask readings for simulating the effect of missing data.

In this paper, we study robust multimodal fusion for the human activity recognition (HAR) task,
assuming people carry one or multiple devices, each equipped with a 9-axis or 6-axis IMU.
These sensors can be heterogeneous and worn on different body parts.
The fusion model should make opportunistic use of the available sensor data, handle time-varying noise as well as continuous blocks of missing data, and achieve high accuracy in HAR by taking advantage of 
the patterns that appear across multiple modalities.
To satisfy these requirements, 
we propose a multimodal fusion model, dubbed \emph{Centaur},
which decouples data cleaning from activity recognition,
such that each objective can be achieved using a lightweight, yet effective machine learning model.
Specifically, we build Centaur's data cleaning module based on 
a denoising autoencoder~(DAE) that employs stacked convolutional layers with large kernels in the encoder 
to mitigate the data quality issues while extracting compressed latent representations. 
These representations, which are largely insensitive to missing and noisy data,
are decoded using transposed convolutional layers to produce a cleaned version of the sensor data.
The activity recognition module of Centaur extracts temporal feature embeddings for every sensor channel using 
a convolution neural network (CNN).
A self-attention mechanism~\cite{vaswani2017attention} is then applied to the feature embeddings 
to exploit the cross-sensor correlations for effective activity recognition.
These two modules are trained independently, 
making it possible to attach another inference module to the data cleaning module,
\textit{e.g.}, for gesture recognition.
The contributions of this paper are as follows:
\begin{itemize}[topsep=0pt]
    \item We propose a modular multimodal fusion model 
    that is robust to both consecutive missing data and significant noise that varies over time.
    We train Centaur using a stochastic data corruption process that simulates realistic sensor faults.
    
    
    \item 
    We run a microbenchmark on each module of Centaur separately, 
    and show that the data cleaning and HAR modules both achieve outstanding performance. 
    Specifically, the proposed convolutional DAE-based data cleaning module is compared
    with $2$ baselines based on more complex neural network architectures, which are capable of removing noise and generating novel samples, under $4$ types of sensor faults. 
    The proposed HAR model is compared with $4$ baselines 
    that are shown to have superior performance over several multimodal fusion models in the HAR task. ~\cite{li2021units}. 


    \item We conduct thorough evaluation of Centaur on three multimodal HAR datasets, namely PAMAP2, OPPORTUNITY, and HHAR, to study its robustness to sensor faults that might occur simultaneously. 
    We further compare Centaur with $2$ state-of-the-art robust fusion baselines 
    that can directly learn from multimodal data with sensor faults. 
    We find that despite these faults, 
    Centaur can effectively clean the sensor data and recognize activities, 
    outperforming the baselines with complex architectures.
\end{itemize}

\section{Related Work}
\label{sec:Related work}

\subsection{Sensor Faults}
Sensor faults and failures are common in IoT devices and sensor networks.
Ni et al.~\cite{Ni09} classify sensor faults from a data-centric perspective
into outliers, spikes, stuck-at faults, and noise faults.
The sensor noise is usually modeled using a time-varying multivariate Gaussian distribution,
which is a convenient assumption~\cite{Ni09}.
Raposo et al.~\cite{Raposo2017} classify faults into internal and external faults. 
Internal faults originate inside the sensor, involving one or multiple physical components, \textit{e.g.,} 
transducer, filter, amplifier, and analog-to-digital converter.
External faults originate outside the sensor. 
They include interference, environmental conditions (rain, dust, \textit{etc.}), and overheating.
As discussed in~\cite{Tawakuli23},
many of these faults, including network connection, battery, and hardware issues,
cause a sequence of successive missing data points
(rather than an isolated missing data point) in one or several channels of a given sensor.
For example, the PAMAP2 dataset~\cite{Reiss2012IntroducingAN}, 
which is one of the three IMU datasets we use in this paper,
contains both isolated and sequence missing data, with the average length of
missing data sequences being 31 milliseconds approximately.

There are several techniques to address sensor faults that cause 
continuous blocks of missing data across one or multiple sensor channels~\cite{Adhikari2022}. 
Assuming missing and complete data have the same distribution, 
Zhang et al.~\cite{zhang2019ssim} build a sequence-to-sequence imputation model 
to fill in missing data sequences of varying lengths 
by incorporating information from earlier and later time steps. 
Chen et~al.~\cite{chen2020hgmf} deal with missing data
by using a graph neural network~(GNN) that captures modality interaction information.
Tran et~al.~\cite{Tran2017MissingMI} impute the missing data by employing
stacked residual autoencoders that model the residual between the original and predicted data.
Yi et al.~\cite{yi2016st} propose a spatiotemporal multi-view-based learning method
to address missing data in geo-sensory time-series. 
The authors integrate empirical statistical models, 
such as inverse distance weighting and simple exponential smoothing, 
with data-driven algorithms, such as user-based and item-based collaborative filtering.
Statistical analysis techniques,
such as mean filling, linear interpolation, 
and multiple imputation via chained equations~(MICE)~\cite{little2019statistical}, 
are other alternatives, although they are generally worse than learning-based imputation techniques.
Nevertheless, all these methods only address isolated or sequence missing data,
and are not capable of denoising the existing data points.

\subsection{Multimodal Fusion}
Multimodal data, from one or multiple sensors, 
can be combined for classification using early, late, and multi-level fusion techniques.
In early fusion~\cite{Chen22}, lower dimensional representations of multimodal data 
are concatenated at the input level of the application model (\textit{e.g.,} the HAR model).
A special case of early fusion is when a shared representation is learned 
for different modalities~\cite{Ngiam11, Jeya2019}.
In late fusion (\textit{aka} decision-level fusion)~\cite{Saha2021}, 
outputs of unimodal application models are aggregated at the end.
Multi-level fusion is a hybrid approach 
where fusion takes place at different stages~\cite{xue2019deepfusion}.

\subsubsection{Incorporating Cross-Sensor Correlation}
Zhang et al.~\cite{zhang2018multimodal} learn common and modality-specific information 
to improve the inference capability of a multimodal emotion recognition model. 
The experiments were performed on audio traces and images to extract useful information. 
As a well-known measure of dependence and a generalized version of Pearson correlation, 
the Hirschfeld-Gebelein-R{\'e}nyi~(HGR) maximal correlation~\cite{renyi1959measures}
is extended in~\cite{wang2019efficient} (named soft-HGR),
to extract informative features from multiple modalities that may contain missing data.


\begin{figure*}[t!]
\centering
\includegraphics[width=1\linewidth]{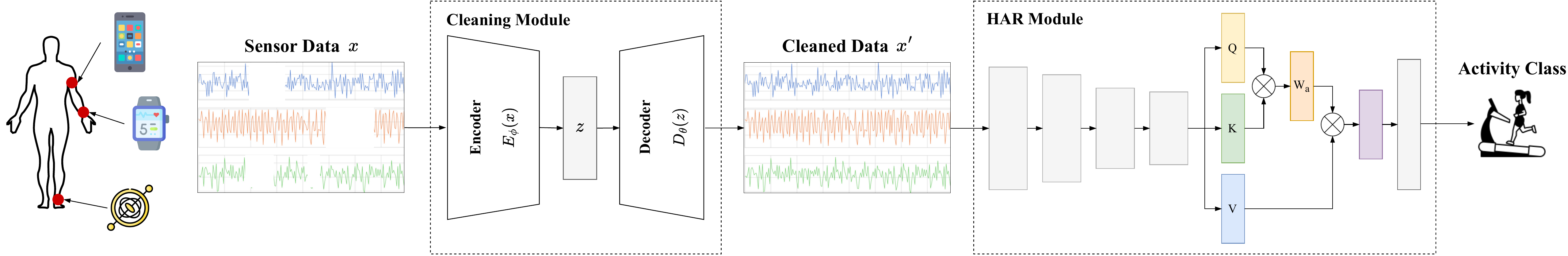}
\vspace{-2mm}
\caption{Overview of Centaur's architecture after cleaning and activity recognition modules are trained independently.}
\label{fig:test-framework}
\end{figure*}

\subsubsection{Multimodal Fusion for HAR}
Deep learning frameworks have been widely adopted for the classification of multimodal data, 
\textit{e.g.,} for activity recognition~\cite{Yao2016, Radu2017, Jeya2019, xue2019deepfusion}.
Radu et al.~\cite{Radu2017} propose feature-concatenated and modality-specific 
deep neural network architectures that use DNN and CNN as base classifiers 
to perform activity and context recognition. 
DeepSense~\cite{Yao2016} integrates a CNN and an RNN to learn dependencies over time and across sensing modalities,
enabling more accurate activity recognition in the presence of noise.
SenseHAR~\cite{Jeya2019} maps raw data collected by the available sensors to a shared low-dimensional latent space, 
representing a virtual sensor that is robust to the availability and variations of the sensors.
The mean of these latent representations is then fed to a pre-trained HAR model to predict the activity label.
DeepFusion~\cite{xue2019deepfusion} fuses readings of multiple sensors using a complex architecture. 
It consists of a sensor-representation (SR) module, 
a weighted-combination (WC) module, and a cross-sensor~(CS) module. 
The SR module consists of multiple CNNs, one for each sensor node,
to learn representations from raw heterogeneous data.
The WC module uses a weighted aggregation strategy to utilize multi-sensor information. 
The correlation between the sensors is captured in the CS module 
by using a single-layer fully connected neural network, 
and the output vector of this module is obtained via averaging. 
Finally, the output of WC and CS modules are concatenated and 
the softmax layer is used to predict the activity label.
STFNets~\cite{yao2019stfnets} introduces a short-time Fourier neural network that can learn frequency domain representations by integrating neural networks and time-frequency analysis. 
A drawback of this model is its complex architecture that relies on multi-resolution layers
which are computationally expensive.
The multi-resolution layers consist of two-dimensional time-series data, 
where each dimension is transformed to the frequency domain at four different resolutions 
using short-time Fourier transform (STFT). 
Similarly, an inverse STFT operation is required to convert the data back to the time domain 
so that it could serve as an input to the next block.
Despite significant advances made toward multimodal fusion,
none of the above papers discusses how the issue of missing data or modalities can be tackled in the HAR task. 

\subsubsection{Robust Fusion for HAR}
Some efforts have been made recently to develop a multimodal fusion model 
that is robust to noisy and incomplete data.
SADeepSense~\cite{Yao2019} is an extension of DeepSense~\cite{Yao2016} 
that introduces a sensor-temporal self-attention module 
to take into account the reliability of heterogeneous sensors.
Experiments were conducted on noise-augmented human activity and gesture recognition datasets. 
Yet, it does not study the effect of missing data. 
Nevertheless, it is possible to substitute missing data with a default value, 
then use SADeepSense to perform HAR, tackling both data quality issues.
UniTS~\cite{li2021units} proposes the short-time Fourier series-inspired neural network, 
named TS-Encoder, and employs multiple TS-Encoders
to extract information in the time and frequency domains at various scales for classification tasks.
Segmenting sensor data using a larger window is more favorable for UniTS 
as it can fully exploit the multi-scale information. 
Besides, the computational overhead increases as more TS-Encoders are adopted.
The authors examine the robustness of the fusion model 
by simulating noisy environments and random missing data. 
For pre-processing, 45 sensor channels and windows of 256 timestamps 
are used to perform a 4-class human locomotion recognition on 
the OPPORTUNITY activity recognition dataset~\cite{Roggen2010}. 
Our work differs from UniTS~\cite{li2021units} in that we consider 
consecutive missing and noisy data simultaneously, 
using a stochastic corruption process described in Section~\ref{subsec:corruption-process},
while they assume each sample may be missing with a probability that is independent of the other samples,
hence long periods of missing data are very unlikely in their work.
Furthermore, our proposed model achieves higher classification accuracy 
and F1~score in the presence of these issues
(see Section~\ref{sec:Evaluation}).

\section{Centaur Architecture}
\label{sec:Centaur Architecture}
\subsection{System Overview}
Centaur is a robust multimodal fusion model that can perform accurate human activity recognition 
given intermittent and noisy data from multiple IMUs.
Figure~\ref{fig:test-framework} depicts its architecture.
Centaur takes as input a 2-dimensional matrix created 
by applying a sliding window to all sensor channels as
described in Section~\ref{sec:Experiment Setup}.
It has two components that are trained independently,
namely a sensor data cleaning module and a human activity recognition module.
The cleaning module leverages a DAE that learns an implicit mapping 
from corrupted sensor data to clean sensor data.
The reconstructed (clean) data will be fed to a novel HAR model 
to perform human activity recognition.

To train the data cleaning module, we simulate sensor faults that are common in real-world settings 
via a stochastic corruption process described in Section~\ref{subsec:corruption-process}.
We use publicly available HAR datasets that are used in prior work as ground truth,
then generate corrupted sensor data by passing data segments sampled from each dataset through the corruption process.
The DAE model takes as input the corrupted data and outputs a clean version 
by optimizing the loss, which is the distance between its output and the ground truth.
This decoupling of cleaning and classification tasks is advantageous
because the HAR module will merely focus on extracting salient features from noise-free and complete data. 
Note that the HAR module is trained independently, without using the stochastic corruption process.


Once these two modules are trained, readings of heterogeneous sensors 
embedded in one or multiple devices are fed into the DAE model, 
bypassing the corruption process which is used only during training.
The reconstructed (clean) data are then fed into the HAR model for activity recognition, 
which is a multi-class classification problem.
We discuss the architecture and training of the cleaning module in Section~\ref{sec:cleaning-module}
and the HAR module in Section~\ref{sec:har-module}.


\subsection{Corruption Process} 
\label{subsec:corruption-process}

Considering the data quality issues that are present in IMU datasets 
(e.g., isolated and sequence missing data in PAMAP2 reported in~\cite{tawakuli2023experience})
and our past experiences with 9-axis IMUs from multiple vendors,
we consider four data corruption modes that cover the following cases: 
sensor data in all channels are perturbed by noise; consecutive data points are missing in some channels;
consecutive data points are missing in all channels of some sensors;
and both noisy and missing data occur in some channels. 
The corruption process can be written in this form: $\tilde{x} = c_i(x,\theta_i)$, 
where $c_i()$ indicates the $i^\text{th}$ corruption mode, 
$x$ refers to the original sensor data which is assumed to be noise-free and complete, 
$\tilde{x}$ is the corrupted sensor data, and $\theta_i$ is the parameter(s) of the respective corruption mode.

\subsubsection*{Mode 1: Random noise is added independently to all channels and time steps}
Data generated by IMUs may contain noise across all channels.
This can be caused by inherent sensor noise, turn-on and in-run biases, 
scale factor and alignment errors~\cite{AnalogDevices},
as well as sensor dislocation and degradation.
While static calibration mitigates some of these issues, 
IMUs must be calibrated dynamically to keep the total error 
within the range specified in their data sheets (which is typically around 1\%),
especially when they are used in a different environment.
In practice, this dynamic calibration is not performed at all times, 
causing the total error to greatly exceed 1\%.
We consider an additive white Gaussian noise, 
\textit{i.e.,} the noise added to each data point (after normalization) is $n \sim N(0,\sigma)$.
To control the amount of noise introduced,
we change the variance of the Gaussian distribution, $\sigma$.
In our experiments, we assume all sensor channels share the same $\sigma$ value,
but $n$ is re-sampled from $N(0,\sigma)$ for each channel and time instant.
The corruption process that generates noisy sensor data can be expressed as follows:
$\tilde{x} = c_1(x, \sigma).$ 


\subsubsection*{Mode 2: Consecutive missing data may appear in all channels independently}
Temporary hardware issues and transitions to a low-energy state
can lead to missing data over intervals of a random length. 
In this corruption mode, we assume such incidents may occur in all sensor channels
and the length of the respective interval, $l_{corr}$, 
follows an exponential distribution, \textit{i.e.,} $l_{corr} \sim \text{Exp}(\lambda_{corr})$. 
Here $\lambda_{corr}$ is the rate parameter of the exponential distribution. 
We define the scale parameter $s_{corr}=1/\lambda_{corr}$ to represent the corruption level;
higher $s_{corr}$ implies corruptions last longer on average. 
Similarly, we assume the interval in which each sensor functions normally, $l_{norm}$, 
follows an exponential distribution, $l_{norm} \sim \text{Exp}(\lambda_{norm})$ 
where $s_{norm} = 1/\lambda_{norm}$ is the corresponding scale parameter.
For a given channel, we assume a normal interval is followed by a missing data interval and vice versa.
The corruption process that generates several consecutive missing data can be expressed as follows:
$\tilde{x} = c_2(x, s_{norm}, s_{corr})$. 
In our experiments, we fix $s_{norm}$ and vary $s_{corr}$ to adjust the corruption level.

\subsubsection*{Mode 3: Consecutive missing data may appear in all channels of some sensors}
This type of error is common when sensors transmit data to a processing node via wireless connections. 
Unstable network connections could result in the loss of consecutive data points from all channels of some sensors.
Mode~3 can be viewed as a special case of Mode~2. 
Hence, we assume the interval that each sensor node functions normally (or is corrupted) 
follows the same exponential distribution as in Mode~2. 
The only difference is that in Mode~2, the normal/corrupted interval is sampled independently for all channels of every sensor;
whereas in Mode~3, all channels of the same sensor experience the same condition, 
so the normal/corrupted interval is sampled for each sensor.
The corruption process of Mode~3 is expressed as: $\tilde{x} = c_3(x, s_{norm}, s_{corr})$. 

\subsubsection*{Mode 4: Both noisy and missing data may appear in all channels}
In the last corruption mode, we consider a challenging case where both noisy data (Mode~1)
and consecutive missing data (Mode~2) can appear simultaneously in different channels.
To simulate this, we first add Gaussian noise to sensor readings in all channels, 
then sample from $\text{Exp}(1/s_{norm})$ and $\text{Exp}(1/s_{corr})$ 
to determine normal and missing data intervals in each channel.
This is equivalent to passing the raw data through $c_1(x,\sigma)$ 
and then passing the result through $c_2(x, s_{norm}, s_{corr})$.
We write this corruption process as:
$\tilde{x} = c_4(x, \sigma, s_{norm}, s_{corr}) = c_2(c_1(x,\sigma), s_{norm}, s_{corr})$. 

\section{Convolutional Denoising Autoencoder for Data Cleaning}
\label{sec:cleaning-module}



\begin{figure*}[!ht]
\centering
\includegraphics[width=1\linewidth]{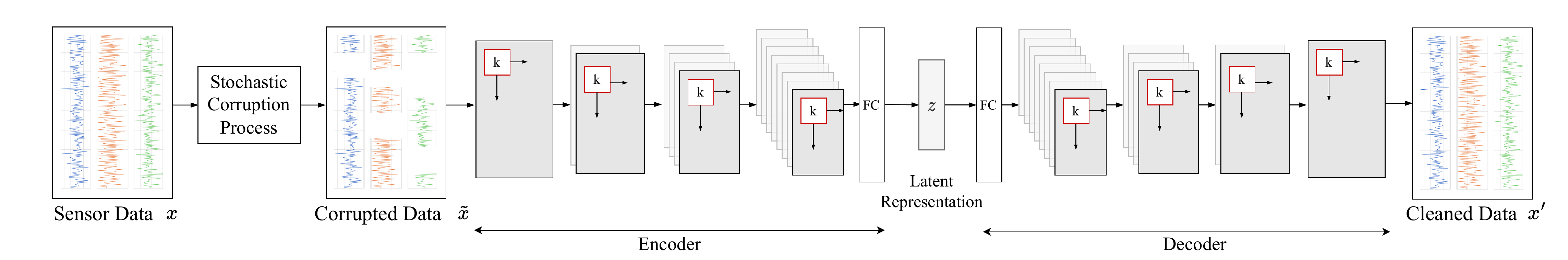}
\vspace{-4mm}
\caption{Architecture of Centaur's data cleaning module. Readings of all sensors ($x$) 
are passed through the corruption process to make the autoencoder 
learn a compressed representation that is useful for reconstructing the data,
and prevent learning a simple identity function.} 
\label{fig:architecture-dae}
\end{figure*}

Centaur's cleaning module is a DAE~\cite{vincent2008extracting} 
comprised of an encoder and a decoder with convolutional layers.
Figure~\ref{fig:architecture-dae} shows the architecture of this DAE 
and the corruption process that we use to train this model.
The DAE learns a mapping from the corrupted sensor readings to the cleaned sensor readings.
To this end, we use a HAR dataset to train the DAE, assuming the original data are noise-free and complete.
The input data is normalized to be in the range of $[0,1]$.
We then pass the original sensor data $x$ through one mode of the corruption process 
as defined in Section~\ref{subsec:corruption-process} to generate the corrupted data $\tilde{x}$, 
which is used as input to the encoder.
We study the data cleaning performance under each corruption mode separately. 
Note that in practice one can either perform model training using the specific corruption mode 
that best matches the real-world setting or use the most general corruption mode (\textit{i.e.}, Mode~4).

The encoder in this DAE learns compressed latent representations 
that are insensitive to various sensor faults that could affect sensor readings.
It contains four stacked two-dimensional convolutional layers (Conv2D), 
each followed by a ReLU activation layer to introduce non-linearity.
In the first convolutional layer, we use $64$ kernels that move with a stride length of $2$ 
to extract feature representations from the corrupted data.
We use a relatively large kernel size of $(5,5)$ such that the receptive field of the convolutional kernel
involves more consecutive data points in more sensor channels. 
This can help learn high-level patterns even when a portion of data is corrupted.
In the next layers, we keep the kernel size the same, 
but double the number of kernels to compensate for the reduced size of the feature map caused by a large kernel.
The feature map from the last convolutional layer is flattened to create a one-dimensional feature vector, 
which is then sent to a fully-connected dense layer to generate the latent representation $z$.

The decoder in this DAE reconstructs a cleaned version of the corrupted sensor data 
given its latent representation $z$. We denote the cleaned data as $x'$.
The decoder has the same architecture as the encoder with one major difference: layers appear in reverse order.
First, a fully-connected dense layer extends the latent representation $z$ such that 
the extended feature vector can be reshaped and processed 
by a transposed two-dimensional convolutional layer (TransposedConv2D). 
We employ four stacked TransposedConv2D layers to recover the convolution process in the encoder.
Each TransposedConv2D layer is followed by a ReLU activation layer.
The number of kernels used in the decoder is the same as the corresponding Conv2D layer in the encoder, 
such that the output of each TransposedConv2D layer in the decoder 
has the same size as the input of the corresponding Conv2D layer in the encoder.
Thus, the output of the last TransposedConv2D layer is the reconstructed version of the input data.
To ensure the range of the reconstructed data is consistent with the normalized input data, \textit{i.e.,} $x' \in [0,1]$, 
we feed the output of the last TransposedConv2D layer to a Sigmoid function 
to obtain the cleaned data $x'$.

\begin{figure}[t]
    \centering
    \begin{subfigure}[b]{0.49\linewidth}
         \centering
         \includegraphics[width=\linewidth]{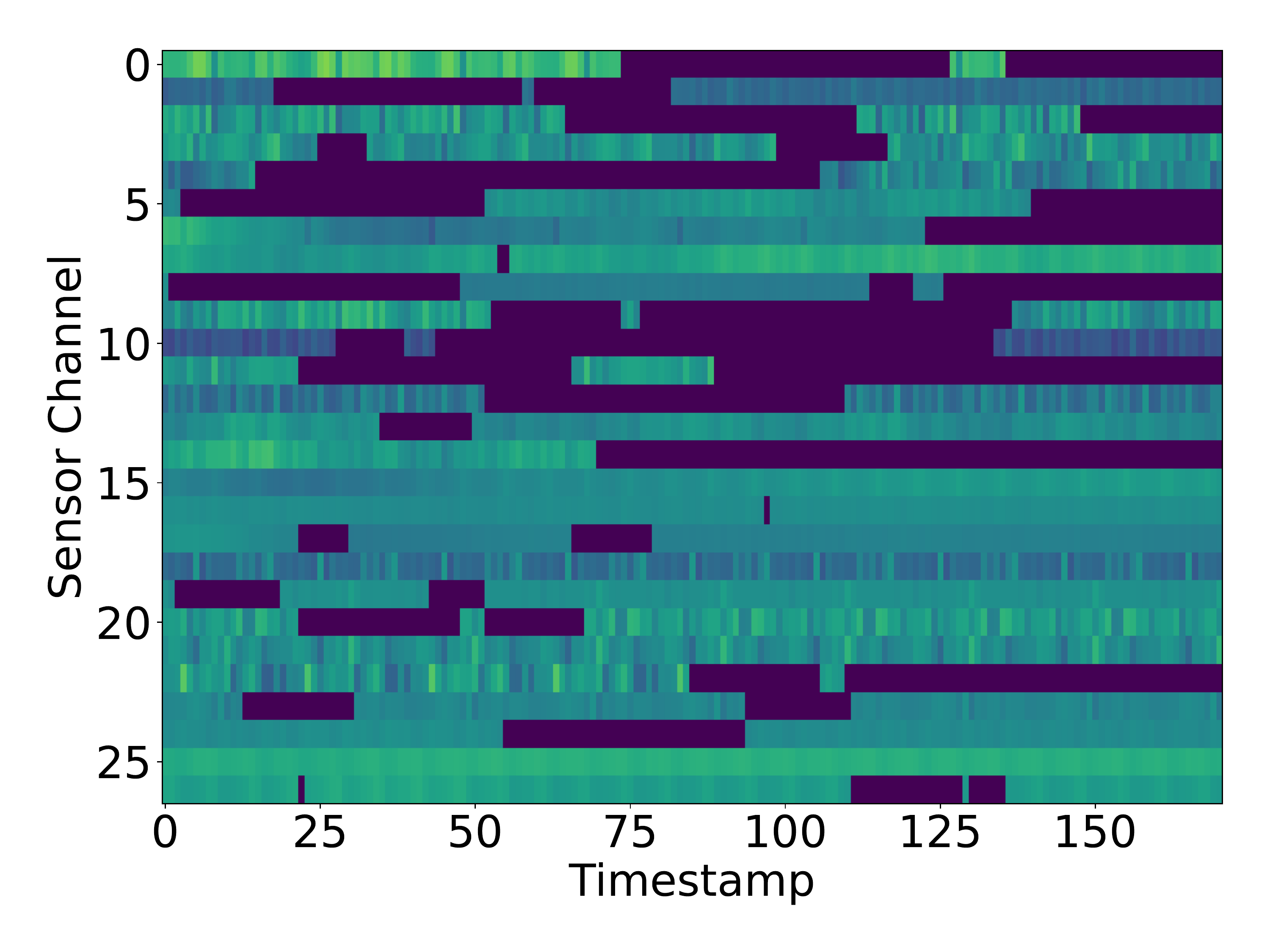}
         \vspace{-7mm}
         \caption{Corrupted data}
     \end{subfigure}
     \hfill
     \begin{subfigure}[b]{0.49\linewidth}
         \centering
         \includegraphics[width=\linewidth]{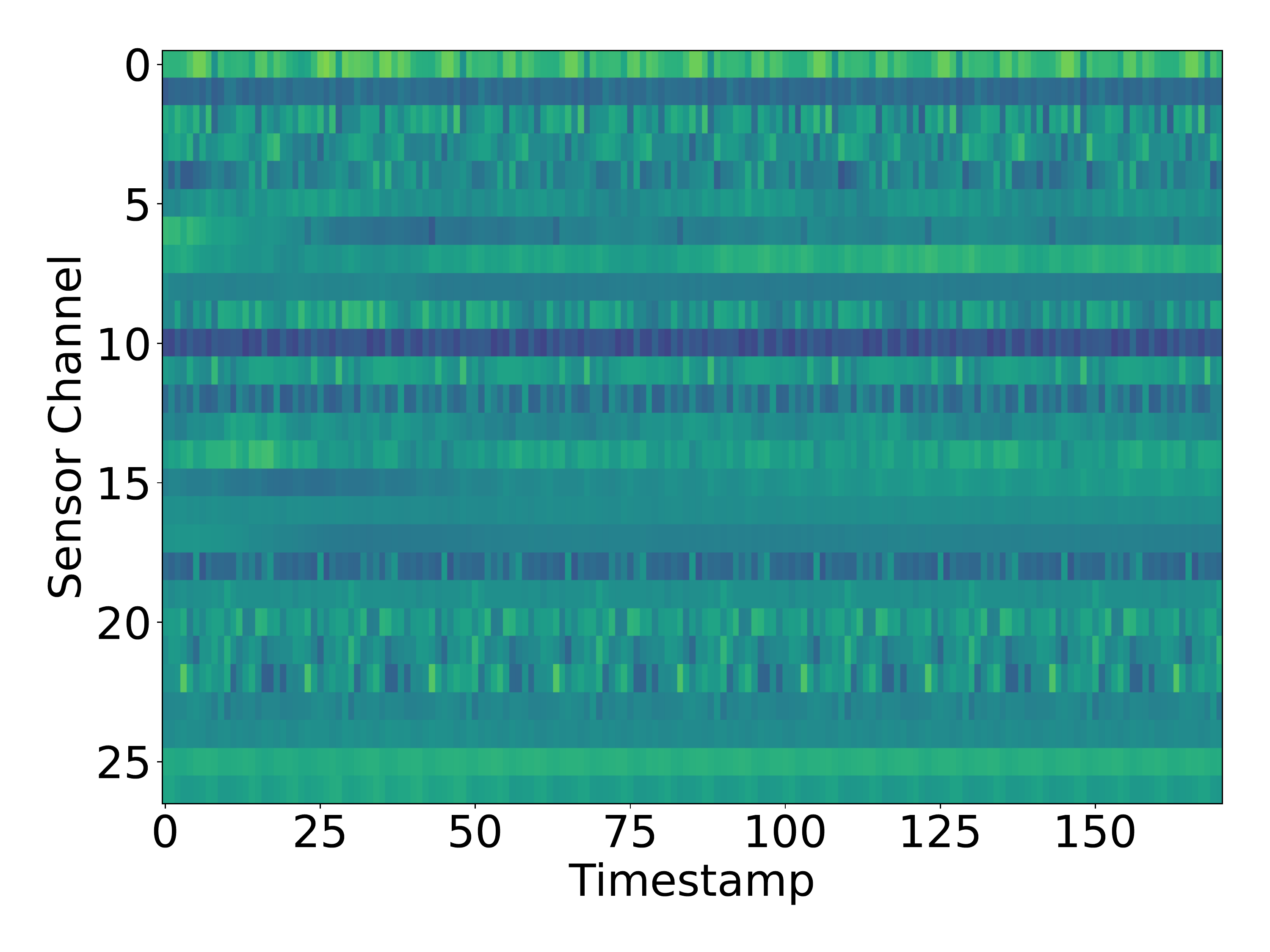}
         \vspace{-7mm}
         \caption{Reconstructed data}
     \end{subfigure}
     \vspace{-2mm}
    \caption{Intermittent sensor data stream (Mode~2) and Centaur's reconstruction.
    Lighter colors show higher values. Missing data points are painted black.}
    \label{fig:recon-sample}
\end{figure}

We use mean squared error~(MSE) between the reconstructed data $x'$ and uncorrupted data $x$
as the loss function of the DAE. 
Assuming the batch size is $N$, the DAE loss can be expressed as:
\begin{equation}
\label{eq:mse_loss}
    \abovedisplayskip=3pt
    \mathcal{L}_{DAE} = \frac{1}{N} \sum_{i=1}^N (x_i'-x_i)^2.
    \belowdisplayskip=3pt
\end{equation}
We use a Root Mean Squared Propagation (RMSprop) optimizer to train the model. 
We empirically set the learning rate to $10^{-4}$ with a momentum of $0.1$.
Figure~\ref{fig:recon-sample} gives an example of how the convolutional denoising autoencoder 
reconstructs the multimodal data that was corrupted using Mode~2.

\begin{figure*}[ht]
\centering
\includegraphics[width=0.8\linewidth]{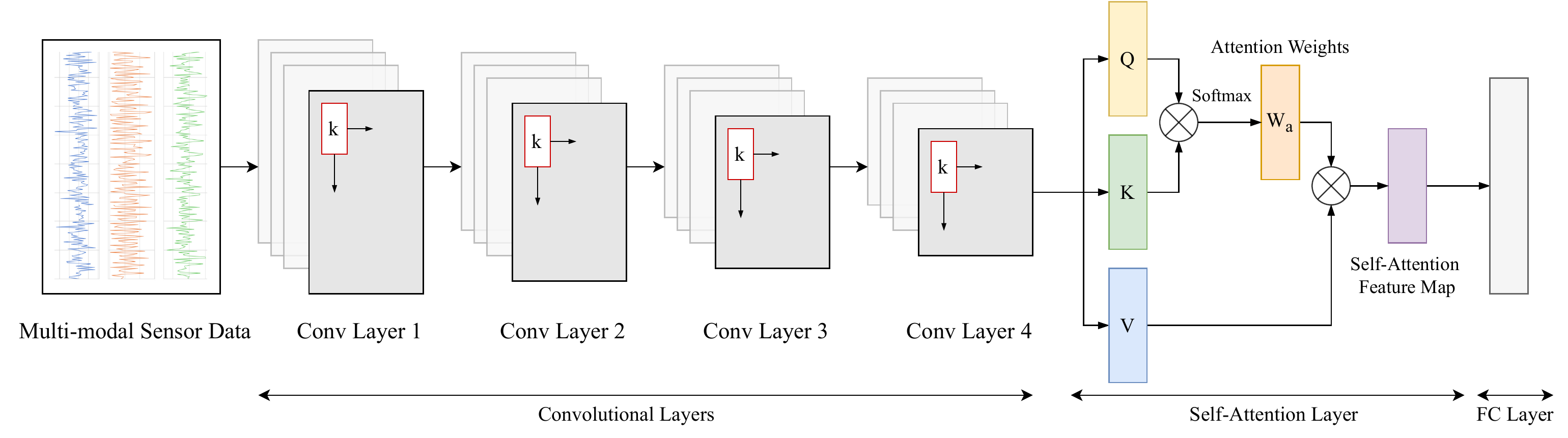}
\vspace{-2mm}
\caption{Architecture of Centaur's HAR model.} 
\label{fig:model-conv-attn}
\end{figure*}

\section{Self-Attention CNN for HAR}
\label{sec:har-module}
We propose a deep neural network that takes advantage of convolutional and attention layers
for multimodal fusion and classification of human activity.
Figure~\ref{fig:model-conv-attn} shows the architecture of the proposed HAR model.
The model contains four stacked convolutional layers serving as a feature extractor 
to obtain compact embeddings of the sensor readings.
A self-attention layer learns from the temporal embeddings to generate a feature map, 
which is flattened and fed to a fully connected layer to predict the activity.

We compose the input of our model by generating a two-dimensional data matrix, 
where the first dimension is the sensor time-series and the second dimension is the available sensor channels.
For each convolutional layer, 
we use $f$ kernels of size $(k, 1)$ to generate the convolutional feature maps, 
where the first dimension, $k$, moves along the time axis to learn and compress features in the temporal domain;
the second dimension, $1$, moves along the sensor channels, such that the rich information 
embedded in each sensor channel is retained in the convolution process 
and cross-channel correlation is learned later in the self-attention layer.
Similar to~\cite{ordonez2016deep}, we do not employ pooling layers in our architecture.
Pooling layers are widely used in CNN-based image recognition tasks to compress the feature space. 
However, when processing multimodal sensor data, 
the number of available sensor channels or the length of the sensor data segments 
can be very limited.
Consequently, applying pooling layers can significantly reduce the available information, degrading the model performance.
Given the input multimodal data segment of size $(H_0 \times W_0)$, 
we can write the feature map size at the $i^\text{th}$ convolutional layer as $(f \times H_i \times W_i)$, 
where $H_i = H_{i-1} - k + 1$, $W_i = W_{i-1} = W_0$, 
$H_0$ is the length of the sliding window, and $W_0$ is the number of available sensor channels.
The four convolutional layers extract the temporal feature representation through multiple kernels 
for each sensor channel.\footnote{We have found empirically that the HAR module performs best with $4$ stacked convolutional layers, 
which is consistent with the observations made in~\cite{ordonez2016deep}.}
We reshape the output of the last convolutional layer
by keeping the temporal dimension $W_4$, 
then flattening the dimension of sensor channels and the number of kernels 
as the input of the following attention layer. 


The convolutional layers learn from the temporal feature representation and 
generate a multi-dimensional feature map for each sensor channel. 
To further exploit the correlations among different sensing modalities and sensor channels, 
we propose to extract such cross-channel correlations 
through the self-attention mechanism~\cite{vaswani2017attention}.
We transform the flattened output of the convolutional layers as the input embedding of the attention layer 
that has a shape of $(H_4 \times (f \times W_4))$, 
where $H_4$ is the compressed temporal dimension that determines the length of the input sequence, 
$f \times W_4$ is the per-channel feature representation extracted by $f$ convolutional kernels.
The query (Q), key (K), and value (V) embeddings are generated using the same input embedding 
to enhance the most significant cross-channel correlations via self-attention weights.
By computing the scaled dot product between Q and K, and passing the results through a softmax activation function, 
we obtain a self-attention weight matrix $W_{a}$ that determines the significance of each feature point:
\begin{equation}
    \abovedisplayskip=3pt
    W_{a} = \text{Softmax}(\frac{Q \cdot K^T}{d_k}),
    \belowdisplayskip=3pt
\end{equation}
where $d_k$ is the dimension of the embedding used to scale the self-attention weights 
to punish large attention weights that would lead to very small gradients.
The self-attention weight is then applied to V to generate the attention feature map.
Lastly, the attention feature map is flattened and fed to a fully connected layer 
to predict the probability of each activity.

We use the cross-entropy loss to train the proposed human activity recognition model.
A stochastic gradient descent (SGD) optimizer is used with a learning rate of 0.01, 
a momentum of 0.9, and a weight decay of $10^{-4}$ to perform model training.

\section{Experimental Setup}
\label{sec:Experiment Setup}
\subsection{Description of HAR Datasets}
We consider a diverse set of HAR datasets that have been previously used in the activity recognition literature. 
Among the publicly available datasets, we select PAMAP2, OPPORTUNITY, and HHAR datasets for evaluation
as they contain the highest number of inertial sensor modalities. 
Note that in each dataset, different values are assigned to $s_{norm}$ and $s_{corr}$ 
according to the length of the time-series segments,
such that we get more than a few normal and missing data intervals in each segment.
Nevertheless, for all three datasets, we choose the value of $\sigma$ from this set $\{0.05, 0.1, 0.2, 0.3\}$.
Below is a brief description of each dataset used in our evaluation.

\subsubsection*{PAMAP2 Physical Activity Monitoring Dataset~\cite{Reiss2012IntroducingAN}}
This dataset consists of 12 different physical activities 
performed by 9 subjects wearing three 9-axis IMUs (accelerometer, gyroscope, and magnetometer). 
The data was sampled at 100Hz and sensor locations are as follows: 
1~IMU sensor over the wrist of the dominant arm, 1~IMU on the chest, and 1~IMU on the dominant side's ankle. 
We follow the preprocessing steps described in~\cite{hammerla2016deep,zeng2018understanding,khaertdinov2021deep} to segment sensor readings using a 5.12~second sliding window with 1s overlap. 
Then the data is down-sampled to 33.3Hz to reduce the computational overhead and normalized between 0 and 1.
Each sliding window contains $171$ data points after down-sampling. 
We set the $s_{norm}$ for corruption Mode~2 and~3 to $80$ and vary $s_{corr}$ 
to be $40$, $50$, and $60$ data points 
to create multiple corruption levels. 
$80\%$ of the data samples are randomly chosen to train the model, and the remaining $20\%$ are used for evaluation.

\subsubsection*{OPPORTUNITY Activity Recognition Dataset~\cite{Roggen2010}}
This dataset covers complex activities performed in a sensor-rich environment. 
We consider the on-body sensors that are mounted on the
left lower arm (LLA), left upper arm (LUA), right lower arm (RLA), right upper arm (RUA), back of the torso, and feet. 
The dataset contains sensor readings for four subjects performing daily activities 
while wearing sensors of different modalities. 
Sensor readings are collected from 5 IMUs, 12 acceleration sensors, and 2 inertial sensors installed on shoes.
Each IMU has $9$ channels, including a 3-axis accelerometer, a 3-axis gyroscope, and a 3-axis magnetometer; 
each acceleration sensor contains 3 axes and each shoe sensor collects data in $16$ channels, 
hence a total of $113$ sensor channels are available. 
The sampling rate of the sensors is 30Hz.
We consider a 5-class locomotion recognition task that involves 4 activities, namely standing, walking, sitting, lying, and a null class.
We follow the preprocessing steps described in~\cite{ordonez2016deep} 
(using a 500ms sliding window) and use the same train-test split.
We set $s_{norm}$ of corruption Mode~2 and~3 to $10$, and $s_{corr}$ to $4$, $6$, and $8$ data points
to create multiple corruption levels.

\subsubsection*{HHAR Heterogeneity Activity Recognition Dataset~\cite{stisen2015smart}}
This dataset contains sensor readings from 9 users performing 6 activities: 
biking, sitting, standing, walking, climbing up the stairs, and climbing down the stairs.
The data is collected by two types of sensors (accelerometer and gyroscope) 
embedded in $8$ smartwatches and $4$ smartphones. 
Each sensor produces readings in $3$ dimensions, hence a total of $6$ sensor channels are available per device.
We only consider the data from smartphones and perform data alignment and sample uniformly separated sensor readings 
following the preprocessing steps described in~\cite{buffelli2021attention}.
We segment the data using $2.5$-second non-overlapping windows, each window containing $100$ data points. 
The data collected from all subjects are mixed together and we randomly draw $80\%$ of the data for training. 
The remaining $20\%$ of the data constitute the test set.
We set $s_{norm}$ of corruption Mode~2 and~3 to $50$, and $s_{corr}$ to $10$, $20$, and $30$ data points 
to obtain multiple corruption levels.

\subsection{Baselines}
We now present the baselines that are used in the next section to evaluate the two modules of Centaur.
\subsubsection{Data Cleaning Baselines}
\subsubsection*{Denoising Adversarial Autoencoder (DAAE)} 
A denoising adversarial autoencoder~\cite{creswell2018denoising} can handle missing and noisy data. 
It uses both denoising and regularization to shape the latent space distribution.
The conditional probability of the encoding given the original data 
can be written as $\tilde{q}_{\phi}(z|x) =  \int q_{\phi}(z|{\tilde{x}}) c (\tilde{x}|x) d \tilde{x}$. 
The conditional probability of the reconstructed data given the encoding is $p_{\theta}(x'|z)$. 
The encoder and decoder are first trained to maximize the likelihood of the reconstructed data 
by minimizing the reconstruction loss $\mathcal{L}_{recon}$.
In our work, we adopt the same MSE loss as introduced in Equation~\ref{eq:mse_loss}.
A discriminator $D_{x}(z)$ is employed to match the conditional probability distribution of latent variables $q_{\phi}(z|\tilde{x})$ to a prior distribution $p(z)$ via adversarial training.
Specifically, the latent feature sampled from the prior distribution $p(z)$ is denoted as $z_{real}$, the latent feature sampled from $q_{\phi}(z|\tilde{x})$ is denoted as $z_{fake}$.
The discriminator aims to differentiate between $z_{real}$ and $z_{fake}$, hence we express the discriminator loss as:
\begin{equation}
    \abovedisplayskip=3pt
    \mathcal{L}_{disc} = - \frac{1}{N} \sum_{i=0}^{N-1}\log D_x(z_{{real}_i}) - \frac{1}{N} \sum_{i=0}^{N-1}\log (1-D_x(z_{{fake}_i})).
    \belowdisplayskip=3pt
\end{equation}
To ensure the latent feature can be sampled from $p(z)$, the decoder is then updated according to the prior loss:
\begin{equation}
    \abovedisplayskip=3pt
    \mathcal{L}_{prior} = \frac{1}{N} \sum_{i=0}^{N-1}\log (1-D_x(z_{{fake}_i})).
    \belowdisplayskip=3pt
\end{equation}
The model training of DAAE involves three steps. First, the encoder and decoder are optimized using $\mathcal{L}_{recon}$. Then, the discriminator is optimized via $\mathcal{L}_{dis}$. Lastly, the decoder is updated via $\mathcal{L}_{prior}$ to match the prior distribution.
 
In our implementation of DAAE, for a fair comparison, we use the same encoder and decoder architecture as in Centaur's data cleaning module.
The discriminator is a 3-layer fully connected network. 
DAAE is trained for 100 epochs with a batch size of 64 using the RMSProp optimizer. 
We set the learning rate as $10^{-4}$ and momentum as $0.1$. 

\subsubsection*{Variational Recurrent Autoencoder (VRAE)}


Unlike Centaur's data cleaning module that relies on convolutional layers to extract latent representations, 
a VRAE~\cite{fabius2014variational} uses recurrent neural networks that are proven effective for learning time-dependent features.
Similar to a DAE, VRAE takes as input the corrupted data $\tilde{x}$ generated by one of the corruption modes 
and outputs the cleaned data $x'$.
The recurrent encoder uses an LSTM layer to generate the hidden state at the current time $h_t$ 
from the hidden state in the past timestamp $h_{t-1}$ and the current sensor reading $\tilde{x}_t$.
The last hidden state $h_{end}$ compresses useful time-domain feature information 
learned from the entire time-series. 
The reparameterization trick originally introduced in~\cite{kingma2013auto} 
is used on $h_{end}$ to sample a latent representation $z$ from the latent space distribution $Z$.

The decoding process generates the clean version of the corrupted input data from the latent feature $z$, 
the probability of which can be written as $p_{\theta}(x'|z)$.
The decoding process is similar to the encoding process, but in reverse order.
First, the sampled latent feature $z$ is decoded into the hidden state representations of the first time stamp $h_1$ using a fully connected layer. 
Then an LSTM layer performs recurrent decoding sequentially.
In the $t^\text{th}$ step of decoding, $h_t$ is fed as input to recover the uncorrupted data 
at the first timestamp, meanwhile generating the hidden state of the next timestamp $h_{t+1}$. 
The loss of VRAE is expressed using the evidence lower bound:
\begin{equation}
    \abovedisplayskip=3pt
    \mathcal{L}_{VRAE} = D_{KL}(q(z|\tilde{x})||p(z)) - \mathbb{E}_{q(z|\tilde{x})}[\log p_{\theta}(x'|z)].
    \belowdisplayskip=3pt
\end{equation}
The first term is the KL~divergence between the true posterior $q(z|\tilde{x})$ and the prior $p(z)$.
The second term is the loss between the reconstructed data and the cleaned data input $x'$.

\subsubsection*{Mean Filling and Linear Interpolation}
Under corruption Mode 2, where we only consider the existence of missing data, 
we further compare our data cleaning module with two widely-adopted data imputation approaches, 
namely mean filling and linear interpolation.
Compared to DAE-based models, mean filling and linear interpolation are model-free and computationally inexpensive.
The comparison helps us understand the improvement that can be made 
by using deep learning-based imputation techniques (although this comes at the cost of slightly increasing the computation overhead).
The mean filling approach imputes the missing periods in each sensor channel using the mean value of 
all the normal data points in that channel.
In linear interpolation, we fit a linear function between the start and end points of the missing data period, 
then uniformly sample the missing values according to this function.

\subsubsection{Human Activity Recognition Baselines}
\subsubsection*{DeepCNN}
We implement a deep convolutional neural network that contains four stacked convolutional layers. The architecture of the DeepCNN baseline is identical to the four convolutional layers in the HAR module of Centaur; 
thus, the comparison between DeepCNN and Centaur's HAR module demonstrates the efficacy of employing the self-attention mechanism.

\subsubsection*{DeepConvLSTM~\cite{ordonez2016deep}} 
It is a deep neural network for HAR that consists of four convolutional layers and two recurrent layers. 
It is capable of automatically learning feature representations and taking into account temporal dependencies. 
The multimodal sensor readings are processed by four convolutional layers 
to compress time domain information and extend the feature dimension for each sensor channel. 
Two LSTM-based recurrent layers extract time-dependent feature representations and 
pass the output through Softmax. 
The design of Centaur's convolutional layers is consistent with DeepConvLSTM, 
where pooling operations are not involved as discussed in Section~\ref{sec:har-module}.
Thus, the comparison between DeepCNNLSTM and Centaur's HAR module gives insight into 
whether the self-attention mechanism can capture cross-sensor correlation 
more effectively than recurrent neural networks.

\subsubsection*{SADeepSense~\cite{Yao2019}}
SADeepSense is an extension of DeepSense~\cite{Yao2016} that 
is designed for robust classification on multi-sensor data. 
The original DeepSense model utilizes stacked CNNs to extract features 
from data segments obtained from multiple sensors, and uses Gated Recurrent Units (GRU) 
to learn temporal dependencies between the feature maps extracted from consecutive time intervals. 
On top of that, SADeepSense integrates an additional Self-Attention (SA) mechanism 
to learn the correlation between different sensors over time. 
The SA module is inserted in the neural network 
where we combine information from the multiple sensors over time.
To use SADeepSense, we first replace missing values in each channel 
with the arithmetic mean of normal data points in that channel.
This is because SADeepSense cannot handle missing values directly.

\subsubsection*{UniTS~\cite{li2021units}}
It is a robust neural network architecture that can learn from multimodal data 
with artificially injected noise and dropped data points. 
The model consists of multiple branches of temporal-spectral encoders (TS-Encoders) 
extracting sensor spectrogram at different scales. 
Each sensor spectrogram fuses frequency-domain information from all available sensors
and passes this information through ResNet for latent feature encoding. 
The multi-scale latent features are fused and fed to a dense layer for the classification task.
In our experiments, we empirically choose the scales based on the length of the sensor data segments.
In~\cite{li2021units}, UniTS is compared with state-of-the-art multimodal fusion models 
that have been developed in recent years and it is shown that it outperforms these models in the HAR task.\footnote{Data quality issues are neglected in this comparison 
as the other models cannot handle missing and noisy data simultaneously.}
Thus, we use it as a baseline for Centaur's HAR module, as well as the whole robust multimodal fusion framework.

\subsection{Evaluation Metrics}
\subsubsection{Metrics for Evaluating Activity Recognition Models}\label{sssec:Metrics for HAR}
We use the activity recognition accuracy and weighted F1~score averaged over $10$ trials
to evaluate HAR models.
We compute the weighted F1 score to better represent the HAR model performance, 
especially when a dataset contains an uneven distribution of human activity classes.
The weighted F1 score is defined as:
\begin{equation}
    \abovedisplayskip=3pt
    F1 = \sum_i \frac{w_i \cdot TP}{TP+\frac{1}{2}(FP+FN)},
    \belowdisplayskip=3pt
\end{equation}
where $w_i$ is the number of samples with activity class $i$ over the total number of samples; 
$TP$, $FP$, and $FN$ stands for true positive, false positive, and false negative, respectively.

\subsubsection{Metrics for Evaluating Data Cleaning Models}
We evaluate the performance of the data cleaning models using two metrics.
The first metric is the root-mean-square error (RMSE) between the original (uncorrupted) sensor data and 
the cleaned data.
An ideal data cleaning module is expected to generate cleaned sensor data with sufficiently low RMSE
as it suggests that the data cleaning module is effective in reducing the distortions caused 
by the corruption process.
Additionally, we use the performance of Centaur's HAR module (accuracy and F1~score) as our second metric.
The HAR model used here is pre-trained on the original (uncorrupted) sensor data.
Hence, higher HAR performance on the data cleaned by Centaur (or a baseline)
implies more effective denoising and imputation for the target task.

\subsubsection{Metrics for Evaluating End-to-end Multimodal Fusion Models}
We evaluate the performance of the end-to-end sensor fusion models by measuring the accuracy and weighted F1 score 
obtained when using the corrupted data to perform activity recognition.
In this case, we cannot use the RMSE metric because robust multimodal fusion models, 
such as UniTS, do not necessarily reconstruct the sensor data in its original format
before they classify the activity.



\subsection{Implementation Details} 
We implement Centaur, DAAE, VRAE, DeepConvLSTM, SADeepSense, and UniTS baselines using PyTorch. 
We followed the PyTorch implementation released by the authors of DAAE~\cite{daae-git}, 
UniTS~\cite{units-git}, and an implementation of VRAE that we found online~\cite{vrae-git}.
We used the Lasagne implementation provided by the authors of DeepConvLSTM as a reference~\cite{deepconvlstm-git} 
and made our best effort to reproduce this work using PyTorch. 
We used our own implementation for SADeepSense.
All models are trained on an NVIDIA RTX 2080 TI GPU.


\section{Evaluation}\label{sec:Evaluation}
We use microbenchmarks to investigate the efficacy of the two modules of Centaur 
before looking at its robustness and performance in comparison with our end-to-end baseline.
For brevity, we only use the OPPORTUNITY dataset in our microbenchmarks, 
and consider all three datasets to evaluate Centaur as a whole. 

\subsection{Human Activity Recognition Module}
Table~\ref{tab:eval-har} shows the performance of the proposed attention-based CNN and HAR baselines on the original (uncorrupted) OPPORTUNITY dataset\footnote{It is worth 
noting that the HAR task is trivial on the uncorrupted PAMAP2 dataset as Centaur's HAR model and all baselines yield an accuracy of around $99\%$.}, 
where the accuracy and weighted F1 score are averaged over $10$ trials.
The DeepCNN baseline, which uses only four stacked convolutional layers, 
yields average accuracy (F1~score) of $86.47\%$ ($86.31\%$).
By incorporating recurrent dense layers, DeepConvLSTM provides a modest improvement in performance,
with average accuracy (F1~score) of $87.05\%$ ($86.61\%$).
Our attention-based model achieves the best HAR performance with average accuracy (F1~score) of $88.78\%$ ($88.69\%$).
SADeepSense yields an average HAR accuracy (F1 score) of $81.01\%$ ($80.23\%$), 
which is lower than the other models. 
We attribute this to its complex architecture that makes it harder to 
generalize to different IMU datasets.
We also evaluate the performance of UniTS on the uncorrupted OPPORTUNITY dataset 
and find that it achieves an accuracy of $86.31\%$ and F1~score of $86.27\%$, 
which is slightly worse than a simple deep convolutional network. 
This relatively poor performance of UniTS might be due to the fact that it requires segmenting sensor data
using a sufficiently large window so as to extract sensor spectrogram on multiple scales. 
However, in the pre-processing step, 
the window length of the OPPORTUNITY dataset is set to $24$ samples 
for a fair comparison with other baselines (in particular DeepConvLSTM).
This is smaller than the $512$ samples used in their work~\cite{li2021units}.
To summarize, our proposed attention-based HAR module outperforms the baselines when there are no data quality issues.
This implies that the proposed architecture is suitable for extracting per-channel temporal features  
in addition to utilizing cross-sensor information for accurate multimodal fusion. 

\begin{table}[t!]
    \centering
    \resizebox{\linewidth}{!}{
    \aboverulesep=0ex
    \belowrulesep=0ex
    \centering
    \begin{tabular}{c|c|c|c|c|c}

    & DeepCNN & DeepConvLSTM & SADeepSense & UniTS  & Centaur  \\
    \toprule
    Accuracy & 86.47 & 87.05 & 81.01 & 86.31 & \textbf{88.78} \\
    \midrule
    Weighted F1 & 86.31 & 86.91 &  80.23 &86.27  &\textbf{88.69} \\
    \bottomrule
   \end{tabular}}
   \vspace{4mm}
     \caption{Human activity recognition performance on the OPPORTUNITY dataset (without noise and missing data).}
     \label{tab:eval-har}
\end{table}

\subsection{Cleaning Module}
\label{subsec:eval-cleaning-module}

\begin{table*}[ht]
    \centering
    \small
    \aboverulesep=0ex
    \belowrulesep=0ex
    \centering
    \resizebox{\linewidth}{!}{
    \begin{tabular}{c|c c c|c c c|c c c|c c c}
    \toprule
    \multicolumn{1}{c|}{\textbf{Corruption}} &
    \multicolumn{3}{c|}{$\sigma$=0.01} &
    \multicolumn{3}{c|}{$\sigma$=0.05} &
    \multicolumn{3}{c|}{$\sigma$=0.1} &
    \multicolumn{3}{c}{$\sigma$=0.2} \\ 
     \textbf{Mode 1} & Accuracy & F1 Score & RMSE & Accuracy & F1 Score & RMSE & Accuracy & F1 Score & RMSE & Accuracy & F1 Score & RMSE  \\ 
    \midrule
    Corrupted Data & 89.15 & 89.06 & 0.01 & 84.55 & 84.77 & 0.05 & 67.87 & 68.41 & 0.1 & 44.86 & 42.58 & 0.2   \\ 
    \midrule 
    DAAE & 86.47 & 86.19 & 0.0300 & 85.65 & 85.31 & 0.0337 & 85.83 & 85.51 & 0.0322 & 83.98 & 83.52 & 0.0352  \\ 
    \midrule
    VRAE & 87.89 & 87.73 & 0.0252 & 87.74 & 87.54 & 0.0259 & 87.12 & 86.90 & 0.0290 & 85.63 & 85.32 & 0.0343  \\ 
    \midrule
    Convolutional DAE & \textbf{88.99} & \textbf{88.88} & \textbf{0.0211} & \textbf{88.49} & \textbf{88.34} & \textbf{0.0233} & \textbf{88.12} & \textbf{87.95} & \textbf{0.0267} & \textbf{87.23} & \textbf{87.01} & \textbf{0.0319}  \\ 
    \bottomrule
   \end{tabular}
  }
    \caption{Data cleaning performance on OPPORTUNITY w/ corruption Mode~1. Accuracy (F1 score) on raw data: $89.21\%$ ($89.12\%$).}
    \label{tab:eval-cleaning-mode1}
\end{table*}

\begin{table*}[ht]
    \centering
    \aboverulesep=0ex
    \belowrulesep=0ex
    \centering
    \small
    \begin{tabular}{c|c c c|c c c|c c c}
    \toprule
    \multicolumn{1}{c|}{\textbf{Corruption}} &
    \multicolumn{3}{c|}{$s_{corr}$=4, $s_{norm}$=10} &
    \multicolumn{3}{c|}{$s_{corr}$=6, $s_{norm}$=10} &
    \multicolumn{3}{c}{$s_{corr}$=8, $s_{norm}$=10}\\
    \textbf{Mode 2}  & Accuracy & F1 Score & RMSE & Accuracy & F1 Score & RMSE & Accuracy & F1 Score & RMSE  \\
    \midrule
    Corrupted Data & 31.58 & 25.24 & 0.2479 & 28.62 & 21.35 & 0.2819 & 27.02 & 19.61 & 0.3033 \\ 
    \midrule
    Linear Interpolation & 76.92 & 77.40 & 0.0659 & 67.08 & 67.77 & 0.0747 & 60.02 & 60.50 & 0.0803 \\ 
    \midrule
    Mean Filling & 88.13 & 87.97 & 0.0220 & 87.46 & 87.27 & 0.0346 & 86.77 & 86.56 & 0.0480 \\ 
    \midrule
    DAAE & 88.59 & 88.47 & 0.0168 & 88.17 & 88.02 & 0.0201 & 88.05 & 87.89 & 0.0224 \\ 
    \midrule
    VRAE & 88.69 & 88.58 & 0.0168 & 88.40 & 88.27 & 0.0199 & 88.30 & 88.16 & 0.0220 \\ 
    \midrule
    Convolutional DAE & \textbf{88.87} & \textbf{88.76} & \textbf{0.0130} & \textbf{88.75} & \textbf{88.64} & \textbf{0.0155} & \textbf{88.68} & \textbf{88.57} & \textbf{0.0172} \\ 
    \bottomrule
   \end{tabular}
     \caption{Data cleaning performance on OPPORTUNITY w/ corruption Mode~2. Accuracy (F1 score) on raw data: $89.21\%$ ($89.12\%$).}
     \label{tab:eval-cleaning-mode2}
\end{table*}

\begin{table*}[ht]
    \centering
    \aboverulesep=0ex
    \belowrulesep=0ex
    \centering
    \small
    \begin{tabular}{c|c c c|c c c|c c c}
    \toprule
    \multicolumn{1}{c|}{\textbf{Corruption}} &
    \multicolumn{3}{c|}{$s_{corr}$=4, $s_{norm}$=10} &
    \multicolumn{3}{c|}{$s_{corr}$=6, $s_{norm}$=10} &
    \multicolumn{3}{c}{$s_{corr}$=8, $s_{norm}$=10}\\
    \textbf{Mode 3}  & Accuracy & F1 Score & RMSE & Accuracy & F1 Score & RMSE & Accuracy & F1 Score & RMSE  \\
    \midrule
    Corrupted Data & 32.87 & 27.49 & 0.2477 & 29.89 & 23.26 & 0.2818 & 28.31 & 21.24 & 0.3034 \\ 
    \midrule
    Linear Interpolation & 75.75 & 76.25 & 0.0659 & 66.68 & 67.34 & 0.0748 & 59.52 & 59.97 & 0.0804 \\ 
    \midrule
    Mean Filling & 87.89 & 87.72 & 0.0219 & 87.06 & 86.87 & 0.0345 & 86.12 & 85.91 & 0.0481 \\ 
    \midrule
    DAAE & 88.31 & 88.18 & 0.0185 & 87.49 & 87.31 & 0.0222 & 86.40 & 86.16 & 0.0259 \\ 
    \midrule
    VRAE & 88.68 & 88.57 & 0.0149 & 88.45 & 88.33 & 0.0176 & 88.38 & 88.24 & 0.0195 \\ 
    \midrule
    Convolutional DAE & \textbf{88.91} & \textbf{88.90} & \textbf{0.0130} & \textbf{88.71} & \textbf{88.60} & \textbf{0.0156} & \textbf{88.74} & \textbf{88.62} & \textbf{0.0173} \\ 
    \bottomrule
   \end{tabular}
     \caption{Data cleaning performance on OPPORTUNITY w/ corruption Mode~3. Accuracy (F1 score) on raw data: $89.21\%$ ($89.12\%$).}
     \label{tab:eval-cleaning-mode3}
\end{table*}

\begin{table*}[ht]
    \centering
    \small
    \aboverulesep=0ex
    \belowrulesep=0ex
    \centering
    \begin{tabular}{c | c c c | c c c | c c c}
    \toprule
    \multicolumn{1}{c|}{\textbf{Corruption}} &
    \multicolumn{3}{c|}{$\sigma$=0.05, $s_{corr}$=4, $s_{norm}$=10} &
    \multicolumn{3}{c|}{$\sigma$=0.1, $s_{corr}$=6, $s_{norm}$=10} &
    \multicolumn{3}{c}{$\sigma$=0.2, $s_{corr}$=8, $s_{norm}$=10}\\
     \textbf{Mode 4} & Accuracy & F1 Score & RMSE & Accuracy & F1 Score & RMSE & Accuracy & F1 Score & RMSE  \\
    \midrule
    Corrupted Data & 30.29 & 23.31 & 0.2520 & 28.36 & 21.67 & 0.2938 & 25.45 & 17.24 & 0.3448 \\ 
    \midrule
    DAAE & 85.45 & 85.13 & 0.0348 & 82.44 & 81.86 & 0.0375 & 76.16 & 74.95 & 0.0447 \\ 
    \midrule
    VRAE & 84.06 & 83.67 & 0.0355 & 81.65 & 81.09 & 0.0389 & 75.66 & 74.39 & 0.0446 \\
    \midrule
    Convolutional DAE & \textbf{88.26} & \textbf{88.10} & \textbf{0.0260} & \textbf{87.70} & \textbf{87.51} & \textbf{0.0302} & \textbf{84.74} & \textbf{84.39} & \textbf{0.0362} \\ 
    \bottomrule
   \end{tabular}
     \caption{Data cleaning performance on OPPORTUNITY w/ corruption Mode~4. Accuracy (F1 score) on raw data: $89.21\%$ ($89.12\%$).}
     \label{tab:eval-cleaning-mode4}
\end{table*}


We evaluate the performance of the three autoencoder-based data cleaning models (Centaur's cleaning module, DAAE, and VRAE)
under the three corruption modes described in Section~\ref{subsec:corruption-process}.
For Mode~2, where we only consider missing data without introducing noise, 
we further compare the autoencoder-based models with mean filling and linear interpolation, 
which are widely adopted data imputation techniques.
We assume the data corruption levels in the training and test phases are identical in these experiments.
Regardless of how data is cleaned, we always pass it through the same attention-based HAR model 
to measure the accuracy and F1~score. 

As shown in Table~\ref{tab:eval-cleaning-mode1}, 
when small white Gaussian noise with $\sigma=0.01$ is introduced, 
feeding the corrupted data to the HAR model directly 
achieves a negligible performance drop compared to when raw data is used, 
yielding an average accuracy (F1~score) of $89.15\%$ ($89.06\%$). 
Cleaning data with none of the data cleaning models can improve the HAR performance.
When slightly increasing the noise level to $\sigma=0.05$, the HAR performance decreases by $\sim5\%$.
Although all three autoencoder-based models improve the HAR performance by denoising the corrupted data, 
DAAE shows the least improvement in accuracy (by $1.10\%$) and weighted F1 score (by $0.54\%$). 
Examining the results for each activity, 
we find that DAAE improves the recognition accuracy of walking by around $16\%$ compared to the corrupted data case, 
but the accuracy of standing and sitting activities drops by around $8\%$ and $17\%$, respectively.
Both VRAE and convolutional DAE show strong performance. 
That said, convolutional DAE, which is our proposed method, outperforms VRAE in terms of accuracy (F1 score) by $0.75\%$ ($0.80\%$).
When increasing the standard deviation of the white Gaussian noise to $\sigma=0.1$ or higher, 
the measurement noise significantly decreases the HAR performance (by more than $16\%$).
DAAE is effective in denoising sensor data, yet it is still worse than VRAE and convolutional DAE.
VRAE shows a more stable performance across different noise levels compared to DAAE, 
yet it is still worse than the proposed convolutional DAE ($1$--$2\%$ lower for all four noise levels).
Convolutional DAE has the best denoising performance among the three autoencoder-based models. 
The above observations based on accuracy and F1 score are consistent with 
the ones that can be made by looking at the RMSE metric, 
where convolutional DAE shows the best denoising capability, 
reducing RMSE by at least $2.14\times$ when $\sigma=0.05$, 
and at most $6.27\times$ when $\sigma=0.2$.

Next we look at Mode 2 of the corruption process.
As it can be seen in Table~\ref{tab:eval-cleaning-mode2}, 
the corrupted data has RMSE of $0.2479$ or higher, and results in F1 score of $30\%$ or lower 
(assuming missing data points are treated as zeros),
even under the lowest corruption level $c_2(x,4,10)$.
Thus, this data is unusable for activity recognition before imputation.
Despite the significant impact of the consecutive missing data,
we observe that all three autoencoder-based models show satisfactory performance by imputing consecutive missing data.
In particular, even DAAE, which underperforms VRAE and DAE, shows an average accuracy (F1 score) of $88.05\%$ ($87.89\%$) under the highest corruption level.
Convolutional DAE has the best performance in this case. 
Compared to uncorrupted data, the HAR accuracy (F1 score) only decreases by $0.53\%$ ($0.55\%$) for $c_2(x,8,10)$. 
Meanwhile, the RMSE is improved by $17.63\times$ compared to the corrupted data case.
Although mean filling demonstrates competitive data imputation capability and has low computation overhead,
the HAR accuracy on the mean-filled data is $1.28\%$ lower than DAAE and $1.91\%$ lower than convolutional DAE under the highest corruption level
$c_2(x,8,10)$.
We observe that linear interpolation is not an effective imputation technique and it becomes worse as the average length of the missing data interval increases.
We believe this is because this imputation technique cannot recover the activity-related patterns that were present in the time-series data.

Looking at the results of Mode~3 in Table~\ref{tab:eval-cleaning-mode3}, 
we observe very similar patterns as in Mode~2.
Specifically, when using the same parameters for $s_{corr}$ and $s_{norm}$, 
directly feeding the data corrupted by Mode~3 yields slightly higher HAR accuracy and F1~score compared to Mode~2. 
Consequently, all three data cleaning approaches can achieve a slightly better HAR accuracy and F1~score under Mode~3. 
Due to the similarity in the corruption level, we only discuss Mode~2 to simulate consecutive missing data 
in the remainder of the paper.

Last but not least, we study the data cleaning performance under the most challenging corruption mode, \textit{i.e.,} Mode~4, 
where both white Gaussian noise and consecutive missing data exist.
We present the result in Table~\ref{tab:eval-cleaning-mode4}.
Convolutional DAE still shows the best performance with the lowest RMSE.
When compared with the HAR accuracy (F1 score) achieved using uncorrupted data, 
it only reduces the performance by 
$0.95\%$ ($1.02\%$), $1.51\%$ ($1.61\%$), and $4.47\%$ ($4.73\%$) for $c_4(x, 0.05, 4, 10)$,  $c_4(x, 0.1, 6, 10)$, and $c_4(x, 0.2, 8, 10)$, respectively. 
Looking at the performance breakdown by activity under the highest corruption level, 
convolutional DAE can bring the classification accuracy of walking, lying, and the null class 
to the same level as obtained using uncorrupted data.
Compared to directly performing HAR using the corrupted data, 
the classification accuracy of the standing activity 
increases by $\sim3.5\%$, and 
the accuracy of the sitting activity increases by nearly $25\%$.
In summary, we have established that Centaur's data cleaning module has the best performance among the three autoencoder-based models. 

\subsection{End-to-end Robust Multimodal Fusion}
We compare the performance of Centaur and two robust fusion baselines, namely UniTS and SADeepSense, 
when they are used for human activity recognition given noisy and incomplete data.
Our evaluation is carried out on the three HAR datasets listed in Section~\ref{sec:Experiment Setup}.
We consider a practical scenario in which the noise variance and average length of the missing data interval are unknown 
when we train the data cleaning module of Centaur. 
Therefore, for each corruption mode, we train Centaur, UniTS, and SADeepSense 
using a corruption process with fixed parameters, and evaluate their performance on different corruption levels.
Specifically, in Mode~1 and~4, we set $\sigma=0.1$ to train the models.
As for $s_{corr}$ and $s_{norm}$ in Mode~2 and~4, we use the values specified for each dataset in Section~\ref{sec:Experiment Setup} 
and train the models using the middle $s_{corr}$ value. 

\begin{figure*}[t!]
\centering
\includegraphics[width=0.8\linewidth]{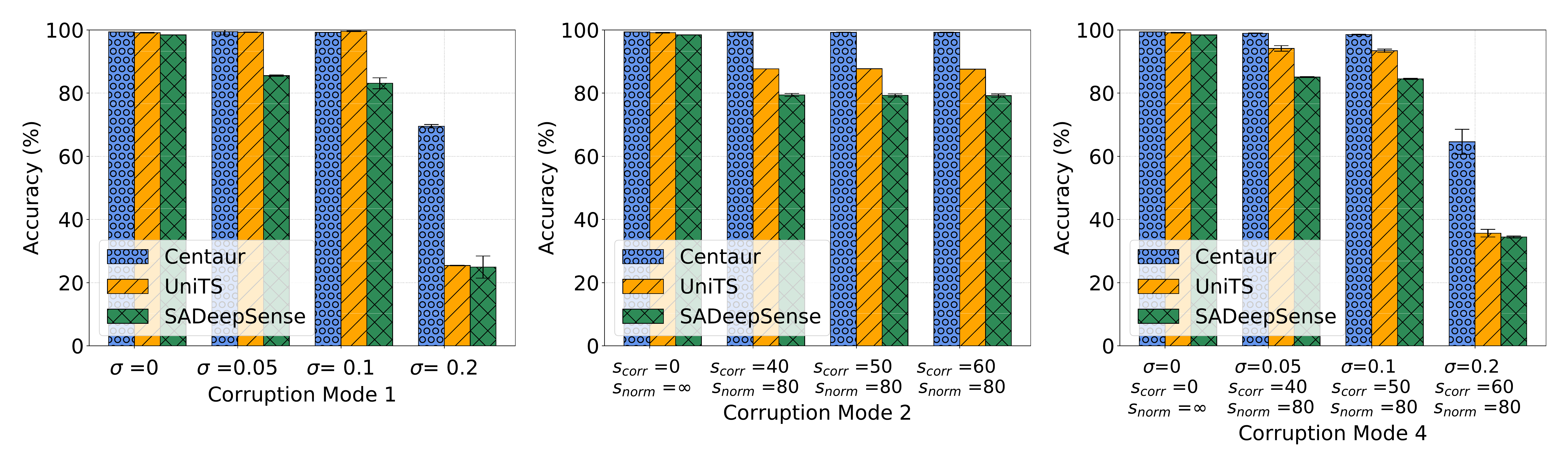}
\vspace{-5mm}
\caption{Performance comparison of the baseline model and Centaur on PAMAP2 dataset. Error bars show the standard deviation across $5$ runs.}
\label{fig:PAMAP2}
\end{figure*}

We first discuss the result (i.e., average accuracy across 5 runs) 
obtained for the three corruption modes (Mode~1, 2, and 4) 
on the PAMAP2 dataset as shown in Figure~\ref{fig:PAMAP2}.
When IMU data is not corrupted, the performance of UniTS and Centaur are almost on par, 
and almost $1\%$ higher than SADeepSense.
After incorporating the white Gaussian noise with $\sigma=0.05$ and $\sigma=0.1$, 
the accuracy of Centaur (and UniTS) remains at the same level of $>99\%$. 
However, the accuracy of SADeepSense decreases significantly to $85.53\%$ when $\sigma=0.05$ and 
to $83.11\%$ when $\sigma=0.1$.
For the highest noise level that we considered, \textit{i.e.}, $\sigma=0.2$, 
Centaur still achieves an average accuracy (F1 score) of $69.52\%$ ($68.43\%$), 
whereas the accuracy of both UniTS and SADeepSense fall down to $\sim 25\%$.
This indicates that Centaur is more robust to high, time-varying noise.
Turning our attention to Mode~2, 
we find that all three robust fusion models show stable performance across all corruption levels, 
with Centaur attaining the highest HAR accuracy of $99.22\%$, 
followed by UniTS achieving an accuracy of $87.6\%$. 
SADeepSense's accuracy is around $8\%$ lower than UniTS.
However, Centaur effectively imputes the consecutive missing data and consistently shows high HAR accuracy for all noise levels. 
In Mode~4, Centaur achieves $98.93\%$ accuracy for the lowest corruption level $c_4(x, 0.05, 40, 80)$, 
while the performance of UniTS (SADeepSense) is $4.80\%$ ($13.86\%$) lower than Centaur. 
When increasing the corruption level to $c_4(x, 0.2, 60, 80)$ -- the most difficult case --
we observe the HAR accuracy of UniTS and SADeepSense decrease by around $58\%$ and $51\%$, respectively, 
whereas Centaur's accuracy only drops by $34\%$. 
Although UniTS successfully learns from data distorted by a noise process with low variance, 
it fails to extract useful information from different modalities 
in the presence of high noise and consecutive missing data.
Despite the complex architecture of SADeepSense, 
it has the worst performance among the three robust fusion models.


\begin{figure*}[t!]
\centering
\includegraphics[width=0.8\linewidth]{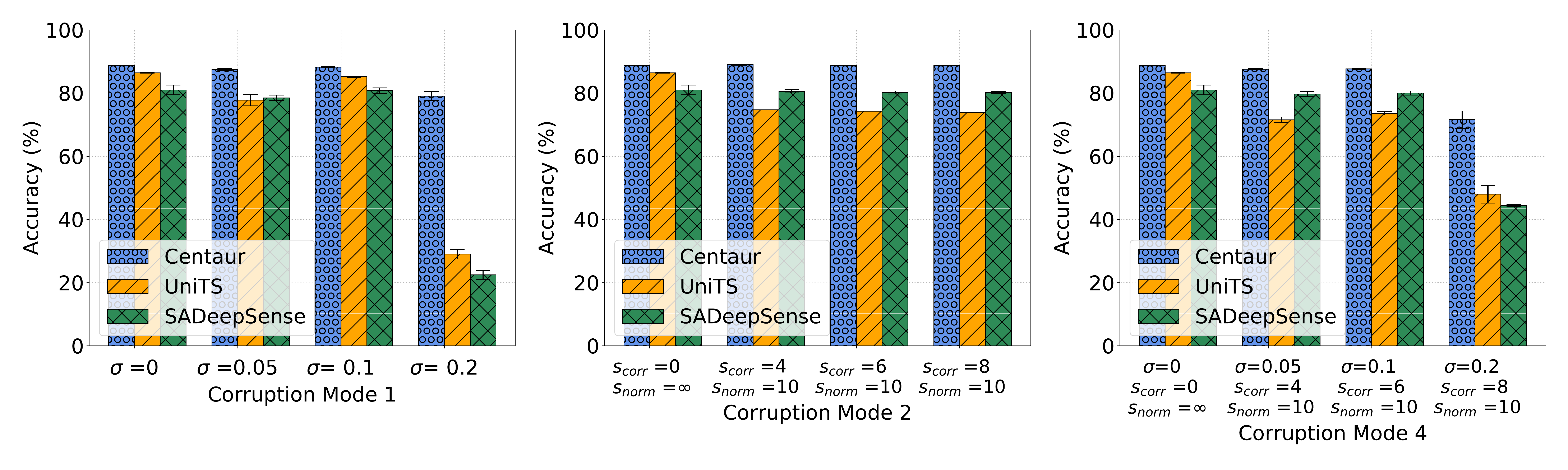}
\vspace{-5mm}
\caption{Performance comparison of the baseline model and Centaur on OPPORTUNITY dataset. Error bars show the standard deviation across $5$ runs.}
\label{fig:OPPORTUNITY}
\end{figure*}

\begin{figure*}[t!]
\centering
\includegraphics[width=0.8\linewidth]{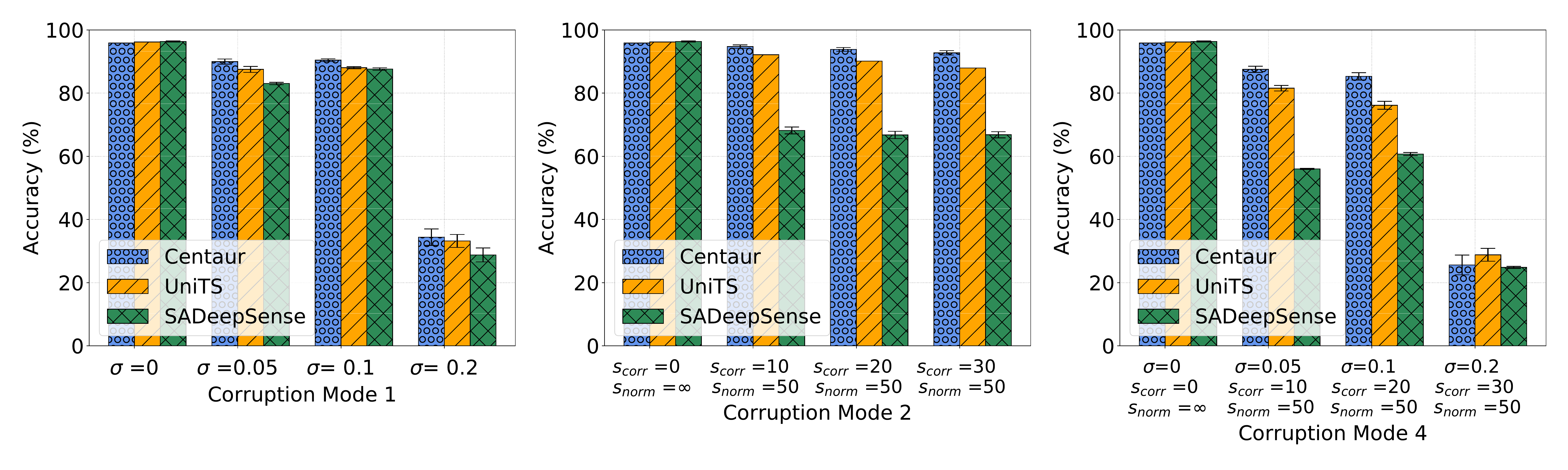}
\vspace{-5mm}
\caption{Performance comparison of the baseline model and Centaur on HHAR dataset. Error bars show the standard deviation across $5$ runs.}
\label{fig:HHAR}
\end{figure*}

Figure~\ref{fig:OPPORTUNITY} compares the end-to-end performance of Centaur and the two baselines on the OPPORTUNITY dataset.
In Mode~1, Gaussian noise does not significantly affect the performance of Centaur when $\sigma$ is smaller than $0.2$. 
UniTS maintains a high HAR accuracy when the corruption level is the same in training and testing (i.e., $\sigma=0.1$) but 
its accuracy drops by $8.66\%$ and $57.38\%$ when the trained model is evaluated on $\sigma=0.05$ and $\sigma=0.2$, respectively, 
indicating slightly worse generalization capability compared to Centaur.
Similar to the PAMAP2 results, SADeepSense performs the worst among the three robust fusion models.
In Mode~2, all three models show consistent performance across all corruption levels. 
Centaur yields over $88\%$ accuracy, followed by SADeepSense which yields over $80\%$ accuracy. 
The accuracy of UniTS is $6\%$ lower than SADeepSense. 
In Mode~4, similar to Mode~2 under the first two corruption levels, 
the three models show similar performance.
Considering a higher corruption level, i.e., $c_4(x,0.2,8,10)$, 
the accuracy of UniTS goes down to $48\%$, which is $23.57\%$ lower than Centaur.
SADeepSense's accuracy is the worst ($44.34\%$) under the highest corruption level.
Overall, we witness that UniTS and SADeepSense can denoise the data under modest noise, \textit{i.e.}, $\sigma=0.05$ and $\sigma=0.1$. 
However, Centaur demonstrates robust fusion capability even with significant noise and longer intervals of missing data. 

Finally, we extend our evaluation to the HHAR dataset and present the result in Figure~\ref{fig:HHAR}.
In this case, the average accuracy of UniTS and SADeepSense are slightly higher than Centaur on uncorrupted data 
although the difference is not statistically significant. 
Specifically, the average accuracy (F1 score) for UniTS, SADeepSense, 
and Centaur are $96.19\%$ ($96.23\%$), $96.34\%$ ($96.26\%$), and $95.84\%$ ($95.89\%$), respectively.
We attribute the better performance of UniTS to the larger number of TS-Encoders (with different receptive scales) 
that can be adopted in the HHAR dataset. 
However, we remark that using more TS-Encoders significantly increases the cost and complexity of training UniTS.
SADeepSense performs well due to its self-attention module that is designed to capture complex dependencies among different sensing inputs over time. 
However, the computational complexity of the self-attention mechanism in SADeepSense 
may be a limiting factor for some resource-constrained devices.
Similar to the above two datasets, we find that both UniTS and Centaur perform well in Mode~2. 
For $c_2(x,30,50)$, Centaur shows an average accuracy (F1~score) of $92.74\%$ ($92.73\%$), outperforming UniTS by $4.80\%$ ($4.82\%$).
However, SADeepSense's HAR accuracy is consistently lower (between $66\%-68\%$) in all cases.
In Mode~1 and Mode~4, all models struggle to denoise the data under the highest corruption level (i.e., $c_1(x,0.2), c_4(x,0.2,30,50)$).
We believe this is due to the small number of sensor channels that are present in the HHAR dataset.
In particular, there are only $6$ sensor channels in the HHAR dataset, 
whereas there is a total of $27$ and $113$ sensor channels in PAMAP2 and OPPORTUNITY, respectively.
This restricts Centaur's ability to take advantage of cross-sensor information. 
Nevertheless, it still outperforms UniTS and SADeepSense 
with respect to accuracy and F1~score under high, time-varying noise.
For example, in Mode~4, Centaur yields an average accuracy (F1 score) of $85.30\%$ ($85.29\%$) for $c_4(x,0.1,20,50)$, 
outperforming UniTS by $9.15\%$ ($9.23\%$). 
SADeepSense performs poorly and only achieves $\sim 60\%$ accuracy.

Overall, the result presented in this section supports the claim 
that Centaur is less expensive computationally
and more robust to consecutive missing data and high noise than the
state-of-the-art multimodal fusion models that tackle these data quality issues. 
When tested on datasets that have multiple sensors with different modalities, 
we found that Centaur captures useful cross-sensor information and takes advantage of it 
to improve the HAR accuracy.

\section{Discussion}
\label{sec:missing_modality}
\begin{figure}[t]
\centering
\includegraphics[width=0.8\linewidth]{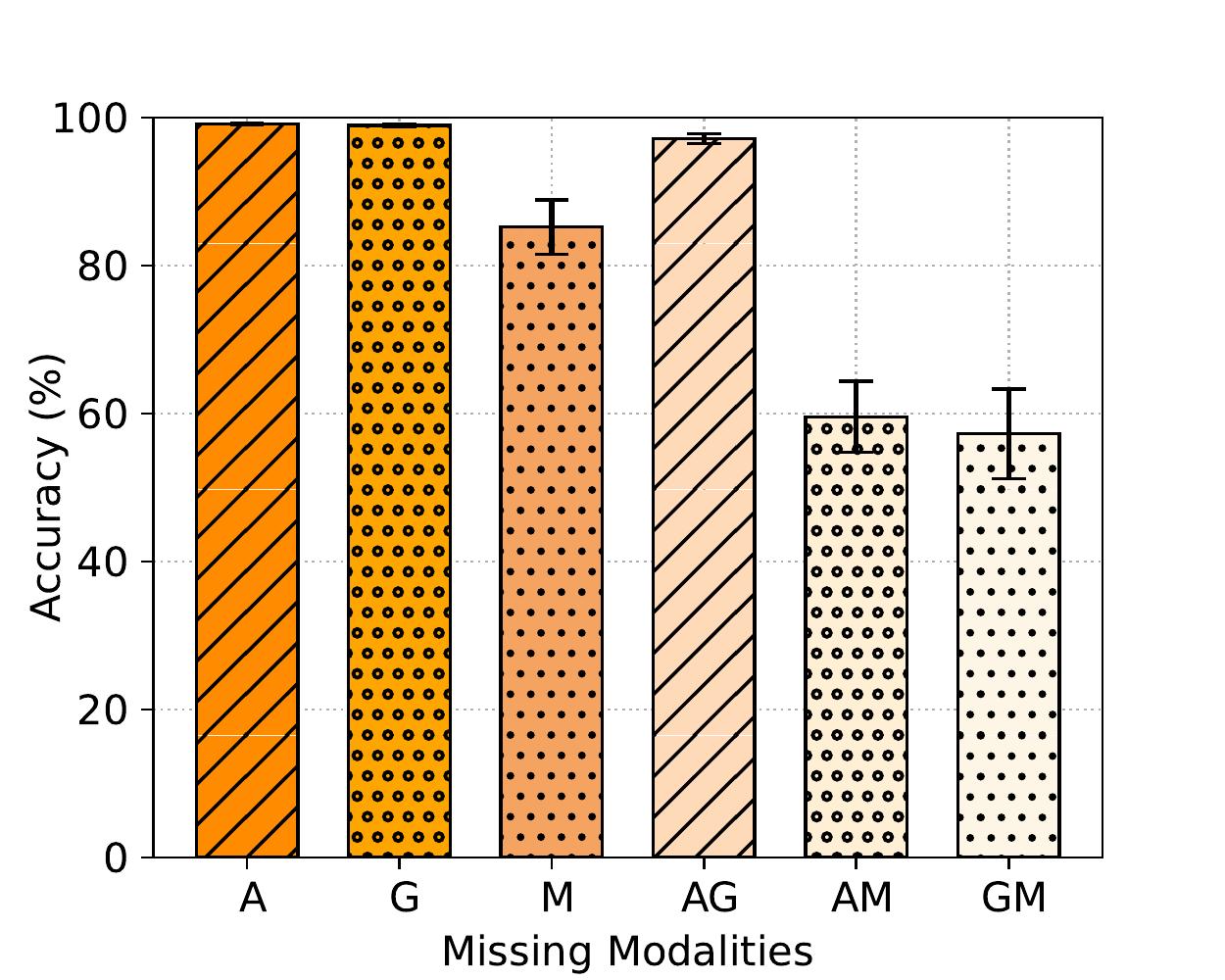}
\vspace{-2mm}
\caption{Illustration of the activity recognition performance of Centaur in case of missing modality.}
\label{fig:missing modalities}
\end{figure}

We discuss a challenging case where data from one or multiple sensing modalities are missing altogether.
Specifically, we assume Centaur is trained using data from 9-axis IMUs
under Mode~4 with $c_4(x, 0.1, 40, 80)$.
We perform our study on the PAMAP2 dataset, where 3 IMUs are involved, 
each containing 3 sensing modalities, namely the accelerometer, gyroscope, and magnetometer. 
We evaluate the performance of Centaur when excluding one or two modalities in all the IMUs, 
and plot the result of three trials in Figure~\ref{fig:missing modalities}.

Centaur treats missing modalities similar to consecutive missing data
and reconstructs the data in a similar fashion.
When a single modality is missing, 
it can be seen that Centaur effectively reconstructs the missing modality (M: magnetometer;
A: accelerometer; G: gyroscope) utilizing information from the two available modalities.
In particular, when data from the accelerometer or gyroscope is missing, 
Centaur achieves more than $98\%$ accuracy and F1 score.
However, when only data from the magnetometer is missing, 
Centaur achieves an average HAR accuracy (F1 score) of $85.23\%$ ($84.61\%$). 
This implies that the information contained in time-series generated by accelerometer and gyroscope in each IMU
is not adequate to recover the magnetometer data.
In any case, a Centaur model that is pre-trained on data from 9-axis IMUs 
can work relatively well on data generated by 6-axis IMUs.

We also study the more challenging case where data from two modalities are missing at the same time. 
This requires Centaur to fully exploit the available information from the remaining modality in each IMU 
to recover data from the other two modalities.
Figure~\ref{fig:missing modalities} shows that when both accelerometer and gyroscope data are missing, 
Centaur can effectively reconstruct both modalities only using the magnetometer data, achieving 
an average HAR accuracy (F1 score) of $97.16\%$ ($97.15\%$). 
This implies that magnetometer data contain rich cross-modality information that can be utilized by Centaur 
to restore both the accelerometer and gyroscope data.
However, when excluding the magnetometer together with accelerometer (gyroscope), 
the average HAR accuracy drops to around $59\%$ ($57\%$),
which suggests that magnetometer data is essential to achieve high accuracy in the HAR task.

\section{Conclusion and Future work} 
\label{sec:Conclusion and Future work}
This paper proposes a neural network architecture for robust multimodal fusion.
We developed a convolutional denoising autoencoder to clean noisy and incomplete sensor data, 
designed three corruption modes to assist with training this model,
and proposed a deep convolutional neural network with the self-attention mechanism 
to perform human activity recognition on the data that is already cleaned.
We showed that Centaur outperforms various baselines and achieves high accuracy 
in the human activity recognition task, despite high noise and large blocks of missing data
that might be the result of hardware, battery, and network issues.
In future work, we plan to study the sensitivity of our result to the sampling rate of sensors,
and explore whether we can unify the sampling rates using a smart combination of resampling and imputation.




\balance

\bibliographystyle{ACM-Reference-Format}
\bibliography{biblo}

\end{document}